\documentclass[times, review, 10pt]{elsarticle}

\usepackage{amssymb}
\usepackage{amsmath}
\usepackage{hyperref}
\usepackage{multirow}
\usepackage{makecell}
\usepackage{booktabs}
\usepackage{caption}
\journal{Pattern Recognition}

\begin{document}


\begin{frontmatter}

\title{Text2Thermal: Physics-Aware Thermal Image Synthesis from Textual Priors}

\author[label1]{Tayeba Qazi\corref{cor1}}
\ead{bsz218186@iitd.ac.in}

\author[label1]{Brejesh Lall}
\ead{brejesh@ee.iitd.ac.in}

\author[label2]{Prerana Mukherjee}
\ead{prerana@jnu.ac.in}

\cortext[cor1]{Corresponding author}

\affiliation[label1]{organization={Bharti School of Telecommunications Technology and Management, Indian Institute of Technology Delhi},
            state={Delhi},
            country={India}}

\affiliation[label2]{organization={School of Engineering, Jawaharlal Nehru University},
            state={Delhi},
            country={India}}

\begin{abstract}
Thermal infrared imaging offers reliable perception in darkness and adverse weather, but
thermal datasets remain scarce, motivating extensive work on translating abundant RGB
images into thermal. Such translation is fundamentally ill-posed as thermal appearance is
governed by surface emissivity and object temperature, neither of which is observable in
the visible spectrum, so a single RGB image is consistent with many valid thermal outputs.
We argue that language offers a natural means of resolving this ambiguity, and propose
Text2Thermal, a framework for physics-aware thermal image synthesis from textual
priors. Rather than inferring the unobservable radiometric factors from RGB, we supply
them explicitly through thermally grounded captions encoding material, weather,
time-of-day, and heat-emission state, and adapt a pretrained Stable Diffusion backbone to
the thermal domain. Because the radiometric content is determined entirely by the prompt,
Text2Thermal synthesizes thermal imagery without requiring a registered RGB image at
inference. Where spatial guidance is desired, an optional control signal imparts scene
geometry without disturbing the prompt-specified radiometry. On M3FD and FLIR, Text2Thermal
achieves state-of-the-art FID among thermal image synthesis methods, and we
additionally report results on the FMB dataset, while offering text-level control that
translation-based approaches cannot provide.
\end{abstract}


\end{frontmatter}

\section{Introduction}
\label{sec:intro}

Text-to-image (T2I) generation has become one of the most active areas of
computer vision. Latent diffusion models trained on web-scale image--text
corpora~\cite{rombach2022high} now synthesise photorealistic imagery from free-form
language, and a rich ecosystem has grown around them, including adapters that inject
spatial control without retraining the backbone~\cite{zhang2023adding,mou2024t2i},
image prompts that supplement text~\cite{ye2023ipadapter}, instruction-based
editing~\cite{Brooks2023InstructPix2Pix}, and attention-level manipulation for
fine-grained semantic control~\cite{hertz2022prompt}. Parameter-efficient
adaptation~\cite{hu2022lora} has made it practical to specialise these models to
new domains at a fraction of the original training cost. Collectively, these
advances have turned the pretrained T2I model into a general-purpose generative
backbone that can be steered toward a target domain rather than retrained for it.

Almost all of this progress has been confined to the visible spectrum.
Text-to-thermal generation, that is, synthesising a thermal infrared (TIR) image
directly from a language description, remains largely unexplored despite the many
applications that depend on thermal images. Because TIR sensing measures emitted
radiation rather than reflected light, it operates independently of illumination and
penetrates fog, smoke and darkness~\cite{vollmer2018infrared,gade2014thermal}. This
makes it valuable for night-time and adverse-weather autonomous
driving~\cite{liu2022target}, pedestrian detection under low light~\cite{llvip}, and
surveillance in conditions where visible sensing fails. Yet thermal datasets remain
small and expensive to collect, because thermal sensors are costly, radiometric
calibration is demanding, and annotating low-texture imagery requires expert effort.
This scarcity is precisely the bottleneck that generative models address in the RGB
domain, where controllable diffusion has been used to synthesise annotated training
data for recognition tasks starved of it~\cite{qin2026diffcrack}, and it motivates
extending large text-conditioned generative models to the thermal modality. Recent
foundation models that simulate the physical world from language and visual
input~\cite{cosmos2025} point the same way, suggesting that such generative priors
need not remain confined to the visible spectrum.

The dominant response to thermal data scarcity has been RGB-to-TIR translation.
Early work adapted conditional~\cite{isola2017image} and
cycle-consistent~\cite{zhu2017unpaired} adversarial translation to the spectral gap,
with ThermalGAN~\cite{kniaz2018thermalgan} specialising the architecture to infrared.
Later methods added edge priors~\cite{lee2023edgeguided} and transformer
backbones~\cite{Chen2024ImplicitMT}, before diffusion models became the dominant
formulation~\cite{Mao2026PID}. These
methods have steadily improved perceptual quality, but they inherit a limitation
that is physical rather than architectural.

A TIR image is a superposition of radiation emitted by an object at its own
temperature, radiation reflected from surrounding objects, and atmospheric
radiation~\cite{vollmer2018infrared,bao2023hadar}, where the emitted term depends on
surface emissivity and absolute temperature, neither of which is observable in the
visible spectrum. A parked car and a car that has just been driven are
indistinguishable in RGB yet appear entirely different in TIR, while two cars of
different colour at the same temperature are indistinguishable in TIR yet clearly
distinct in RGB. The RGB-to-TIR mapping is therefore many-to-many and fundamentally
ill-posed, and a model trained to produce a single deterministic output must
arbitrarily resolve this ambiguity.

We argue that language is the natural vehicle for resolving this ambiguity. The
factors that govern thermal appearance, namely material composition, time of day,
weather, and whether an object is actively emitting heat, are exactly the kind of
scene attributes that text describes well and that the visible spectrum does not
encode. Rather than inferring these factors from RGB, we supply them explicitly as a
structured prior carried through the text channel. Recent work has begun to move in this direction, with TherA~\cite{Lee2026TherA}
conditioning translation on thermally aware descriptions and
T-CLIP~\cite{qazi2026tclipenablingthermalperception} demonstrating text-only thermal
generation as a proof of concept, but a systematic framework for physics-aware
text-to-thermal synthesis has yet to be established. The captions themselves are
derived from paired data following~\cite{Lee2026TherA}, but once trained, the model
requires only text at inference, whereas translation methods need a registered
visible frame for every image they generate.

Building on this observation, we present \textit{Text2Thermal}, a framework for both
unconditional and spatially conditioned text-to-thermal synthesis, where
unconditional denotes the absence of any spatial input. We adapt a pretrained stable diffusion
backbone~\cite{rombach2022high} to the thermal domain through low-rank
adaptation~\cite{hu2022lora}, conditioning it on structured, physics-aware captions
that encode material, weather, time-of-day, and heat-emission state. This transfers
the semantic knowledge of large-scale visible pretraining to a modality for which no
comparable corpus exists, while letting the prompt carry the thermal physics that RGB
cannot. Where paired spatial information is available, we additionally attach a
control branch~\cite{zhang2023adding} that supplies scene geometry, enabling
high-fidelity synthesis aligned to a known layout, which matters when the generated
thermal image is to be paired with annotations drawn for detection or segmentation tasks. We evaluate Text2Thermal on three public
benchmarks, M3FD~\cite{liu2022target}, FLIR~\cite{flir}, and FMB~\cite{liu2023multi},
comparing against both general-purpose image-to-image translation methods and
specialised RGB-to-TIR models.

Our contributions are as follows:
\begin{itemize}
    \item We present a framework for physics-aware text-to-thermal image synthesis,
and show that language conditioning constrains the ambiguity inherent in the
visible-to-thermal mapping.
    \item We adapt a large pretrained text-to-image diffusion model to the thermal
domain using physics-aware captions and parameter-efficient finetuning, achieving the
best reported FID among thermal image synthesis methods on M3FD and FLIR.
    \item We extend the framework with a spatial control branch for structure-aligned
synthesis, and show that the RGB frame provides the strongest structural
conditioning, outperforming other spatial modalities.
    \item We introduce a re-captioning evaluation protocol, comparing captions of
generated and ground-truth thermal images with BERTScore, and use it to show that the
synthesised imagery retains the attributes specified in the prompt.
    \item A leave-one-attribute-out ablation shows that every field of the caption
schema contributes thermal information, with material among the most costly to
remove, consistent with its role in setting emissivity.
\end{itemize}

\section{Related work}
\label{sec:related_work}

\subsection{Thermal image synthesis}

Simulation-based methods model infrared emission explicitly from geometry, material
emissivity, and environmental conditions~\cite{guissin2005irisim,madan2023thermalsynth}.
They are physically grounded by construction, but each scene must be authored manually,
which prevents scaling to the diversity required for training perception models.

Data-driven methods instead learn the visible-to-infrared mapping from paired data.
Early adversarial
approaches~\cite{kniaz2018thermalgan,ozkanouglu2022infragan} were
subsequently constrained with auxiliary priors such as edge
consistency~\cite{lee2023edgeguided} or transformer
backbones~\cite{Chen2024ImplicitMT}, though GAN-based training remains prone to
instability and mode collapse. Diffusion models have since become the dominant
paradigm, with conditioning of increasing sophistication:
ThermalDiff~\cite{qazi2025thermaldiff} performs RGB-conditioned generation,
DiffV2IR~\cite{ran2025diffv2ir} and F-ViTA~\cite{Nair2025FViTA} add segmentation priors
from foundation models~\cite{ren2024grounded}, ThermalGen~\cite{Xu2025ThermalGen}
conditions on dataset-level scene indices, and PID~\cite{Mao2026PID} introduces
physics-based losses derived from a radiometric decomposition of the infrared signal.

All of these assume that a visible image is available at inference and sufficient to
determine the thermal output. That assumption is physically unsound. A thermal image
superposes radiation emitted by an object at its own temperature, radiation reflected
from its surroundings, and atmospheric radiation~\cite{vollmer2018infrared,bao2023hadar},
and the emitted component depends on surface emissivity and absolute temperature,
neither of which is recoverable from reflected visible light. We therefore treat thermal
generation as a text-conditioned synthesis problem, supplying the unobservable factors
through the thermally grounded prompt.

\subsection{Text-to-image synthesis}

Text-conditioned diffusion has advanced rapidly, with latent
diffusion~\cite{rombach2022high} and SDXL~\cite{podell2023sdxl} establishing the
prevailing design: cross-attention between the denoising UNet and a text embedding
produced by a CLIP encoder~\cite{clip}. Because retraining such models for a new
domain is prohibitively expensive, parameter-efficient adaptation has become
standard: LoRA~\cite{hu2022lora} injects trainable low-rank updates into frozen
attention layers.

This body of work is confined almost entirely to the visible spectrum, and applying it
to thermal imagery raises two obstacles. No web-scale thermal-caption corpus exists to
train from, and the text encoder's priors, learned from RGB images, carry no notion of
how a scene radiates heat. We address the first through parameter-efficient adaptation
of a pretrained backbone and the second through captions describing thermally relevant
attributes rather than visible appearance.

\subsection{Spatial control for diffusion models}

Text specifies semantics but not layout, motivating work on injecting spatial
conditions into pretrained generators. ControlNet~\cite{zhang2023adding} clones the
UNet encoder into a trainable branch that accepts an auxiliary map, such as edges,
depth, or segmentation, and adds its features back into the frozen backbone.
T2I-Adapter~\cite{mou2024t2i} achieves comparable control with a lighter module, and
IP-Adapter~\cite{ye2023ipadapter} extends conditioning to image prompts.
Related approaches modulate cross-attention directly~\cite{hertz2022prompt} or follow
editing instructions~\cite{Brooks2023InstructPix2Pix}.

Within the thermal domain, conditional generation has been applied to closing the
synthetic-to-real gap in infrared segmentation~\cite{mayr2024narrowing}, to
object-centric synthesis~\cite{vo2024approach}, and to identity-conditioned infrared
person generation for re-identification~\cite{yu2025identity}, but evaluation in all
three cases is limited to narrow scenarios, and which spatial modality best serves
thermal generation remains open. We examine this question directly.

\subsection{Thermal-aware captions}

The value of text conditioning depends entirely on what the captions encode. Generic
captioning models describe visible appearance and say nothing about how a scene
radiates heat, and thermal benchmarks rarely carry text at all, being built for
detection and segmentation. Useful conditioning therefore requires captions grounded in
the factors that govern infrared emission, namely material emissivity, ambient
conditions, and the heat-emission state of individual objects.

Two recent efforts construct such captions. IR-Cap~\cite{qazi2026tclipenablingthermalperception} addresses this by
conditioning solely on the paired visible frame and issuing two complementary
prompts, where the first elicits scene-level context such as illumination, weather, and
material composition, and the second elicits object-level heat signatures and relative
temperature relationships. TherA~\cite{Lee2026TherA} instead prompts on the visible
and thermal frames jointly, so that the thermal image supplies direct emission
evidence while the visible image anchors object identity and layout, and
canonicalises the result into a schema over scene context, object identity,
material, and heat-emission state. Because its captions are grounded in observed
rather than inferred thermal appearance, we adopt the schema
of~\cite{Lee2026TherA} as our language conditioning, and analyse in
Section~\ref{sec:experiments} which of its attribute classes carry the most thermal
information for generation.
\section{Methodology}
\label{sec:methodology}
\begin{figure}[t]
\centering
\includegraphics[width=\linewidth]{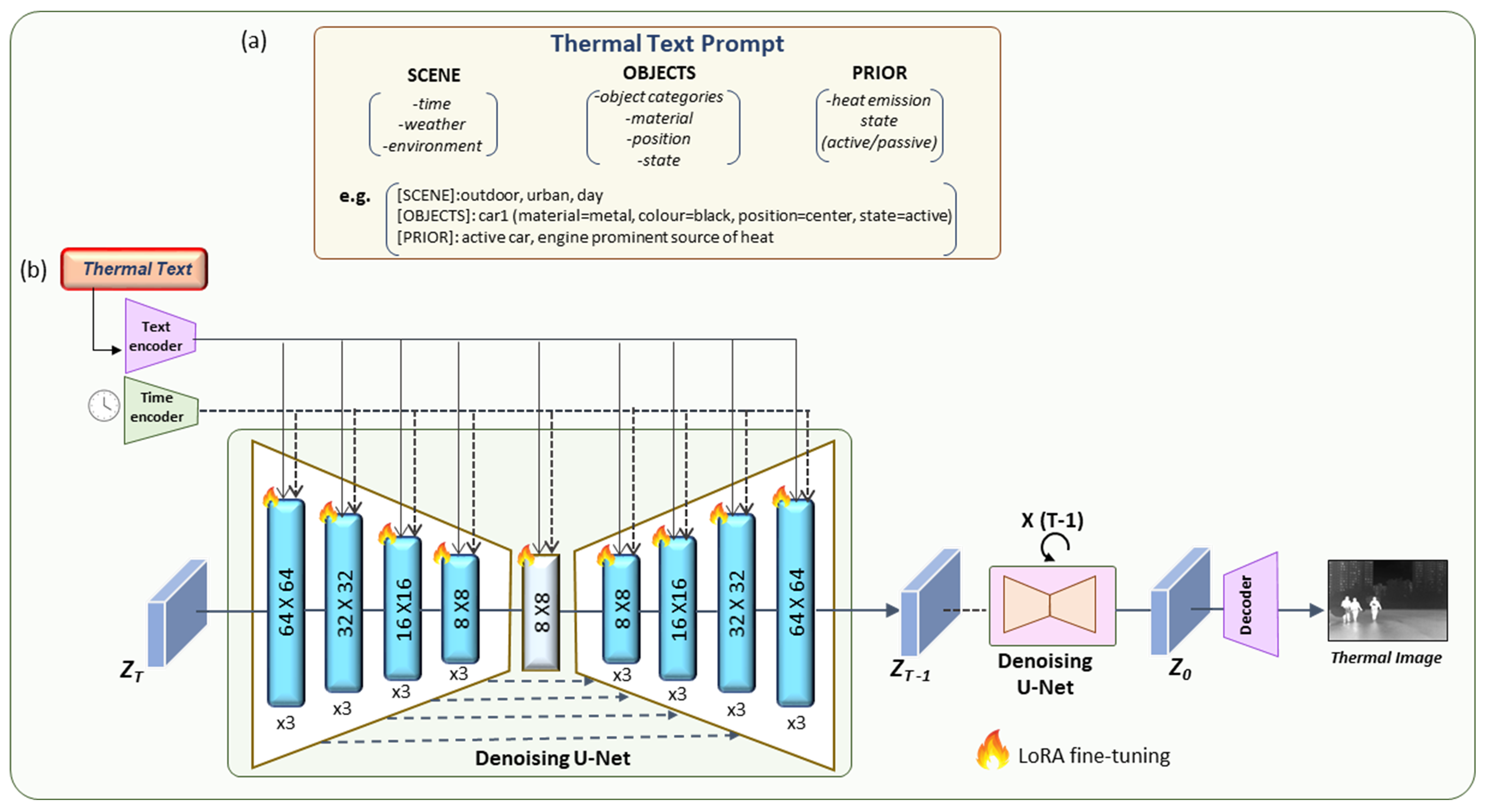}
\caption{Overview of the unconditional Text2Thermal framework. (a) Thermal
physics-aware text prompts describing the fields relevant to thermal
properties. (b) A pretrained stable diffusion model is adapted to the
thermal domain via LoRA fine-tuning of the denoising U-Net, conditioned
jointly on the thermal physics-aware structured caption and a time encoding,
to generate the corresponding thermal image.}\label{fig:sd}
\end{figure}

\begin{figure}[t]
\centering
\includegraphics[width=\linewidth]{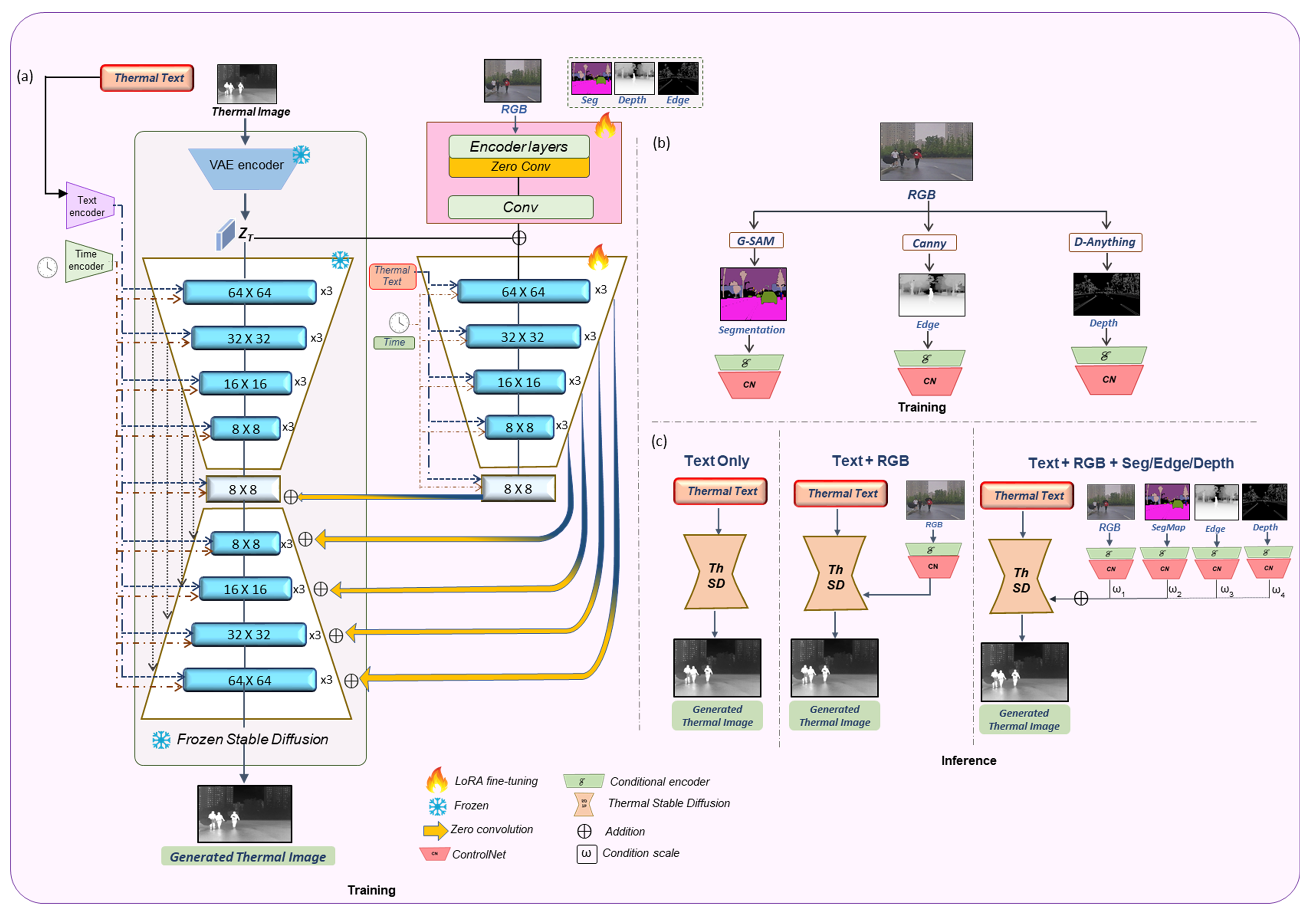}
\caption{Overview of the conditional Text2Thermal framework. (a) Training pipeline: The unconditional Text2Thermal model is kept
frozen, and a control branch initialised from its weights injects an RGB
condition to guide scene structure during thermal image synthesis. (b) Conditional
encoder training: segmentation, edge, and depth maps are extracted from the
RGB frame using G-SAM, Canny, and Depth-Anything respectively, and each
modality-specific conditional encoder is trained
independently. (c) Inference configurations. Text-only; text with RGB; and text
with RGB plus one or more of the segmentation, edge, and depth conditions,
applied with weights $\omega_1$--$\omega_4$.}\label{fig:cn}
\end{figure}
\subsection{Preliminary}
\label{sec:preliminary}

\subsubsection{Stable Diffusion}
\label{sec:stable_diffusion}

Stable Diffusion~\cite{rombach2022high} is a text-conditioned latent diffusion
model. Rather than diffusing in pixel space, it first compresses an image
$\boldsymbol{x}_0 \in \mathbb{R}^{H \times W \times C}$ into a latent
representation using a pretrained autoencoder,
\begin{equation}
\boldsymbol{z}_0 = \mathcal{E}(\boldsymbol{x}_0) \in \mathbb{R}^{h \times w \times c},
\qquad
\tilde{\boldsymbol{x}}_0 = \mathcal{D}(\boldsymbol{z}_0)
\label{eq:sd_vae}
\end{equation}
where the encoder $\mathcal{E}$ downsamples by a factor $f = H/h = W/w$ and the
decoder $\mathcal{D}$ reconstructs the image.

The forward process is a Markov chain that progressively corrupts
$\boldsymbol{z}_0$ with Gaussian noise over $T$ steps,
\begin{equation}
q(\boldsymbol{z}_t \mid \boldsymbol{z}_{t-1}) =
\mathcal{N}\!\left(\boldsymbol{z}_t;\, \sqrt{\alpha_t}\,\boldsymbol{z}_{t-1},\,
(1-\alpha_t)\mathbf{I}\right)
\label{eq:sd_forward}
\end{equation}
with noise schedule $\alpha_t \in (0,1)$. Marginalising the intermediate steps
admits a closed form that samples any timestep directly,
\begin{equation}
\boldsymbol{z}_t = \sqrt{\bar{\alpha}_t}\,\boldsymbol{z}_0
+ \sqrt{1-\bar{\alpha}_t}\,\boldsymbol{\epsilon},
\qquad
\bar{\alpha}_t = \prod\nolimits_{i=1}^{t}\alpha_i
\label{eq:sd_marginal}
\end{equation}
where $\boldsymbol{\epsilon} \sim \mathcal{N}(\mathbf{0}, \mathbf{I})$.

The reverse process is parameterised by a denoising UNet
$\boldsymbol{\epsilon}_\theta$ that predicts the noise added at each step. A text
prompt $y$ is encoded by a frozen CLIP text encoder~\cite{clip} into a token
sequence $\tau(y)$, which is injected into the UNet through cross-attention
layers,
\begin{equation}
\mathbf{Q} = \mathbf{W}_Q\,\varphi(\boldsymbol{z}_t),\quad
\mathbf{K} = \mathbf{W}_K\,\tau(y),\quad
\mathbf{V} = \mathbf{W}_V\,\tau(y)
\label{eq:sd_qkv}
\end{equation}
\begin{equation}
\mathrm{Attention}(\mathbf{Q},\mathbf{K},\mathbf{V}) =
\mathrm{softmax}\!\left(\frac{\mathbf{Q}\mathbf{K}^{\!\top}}{\sqrt{d}}\right)\mathbf{V}
\label{eq:sd_attn}
\end{equation}
where $\varphi(\cdot)$ denotes the intermediate UNet activation and
$\mathbf{W}_Q, \mathbf{W}_K, \mathbf{W}_V$ are learned projections. The model is
trained with the noise-prediction objective
\begin{equation}
\mathcal{L}_{\mathrm{LDM}} =
\mathbb{E}_{\boldsymbol{z}_0, y, t, \boldsymbol{\epsilon} \sim \mathcal{N}(\mathbf{0},\mathbf{I})}
\left[\left\| \boldsymbol{\epsilon} -
\boldsymbol{\epsilon}_\theta(\boldsymbol{z}_t, t, \tau(y)) \right\|_2^2\right]
\label{eq:sd_loss}
\end{equation}

At inference, sampling begins from $\boldsymbol{z}_T \sim \mathcal{N}(\mathbf{0},
\mathbf{I})$ and iteratively denoises to $\boldsymbol{z}_0$, which is decoded to
an image. Prompt adherence is strengthened by classifier-free
guidance~\cite{ho2022classifierfree},
\begin{equation}
\tilde{\boldsymbol{\epsilon}}_\theta =
\boldsymbol{\epsilon}_\theta(\boldsymbol{z}_t, t, \varnothing)
+ s\left(\boldsymbol{\epsilon}_\theta(\boldsymbol{z}_t, t, \tau(y))
- \boldsymbol{\epsilon}_\theta(\boldsymbol{z}_t, t, \varnothing)\right)
\label{eq:cfg}
\end{equation}
where $\varnothing$ denotes the null prompt and $s$ the guidance scale.

\subsubsection{ControlNet}
\label{sec:controlnet}

Text alone specifies semantics but not spatial layout. ControlNet~\cite{zhang2023adding}
addresses this by attaching a trainable branch to a frozen diffusion backbone,
allowing an auxiliary spatial signal to steer generation without disturbing the
pretrained weights.

Let $\mathcal{F}(\cdot\,; \Theta)$ denote a neural block of the backbone that maps
a feature map $\boldsymbol{x}$ to $\boldsymbol{y} = \mathcal{F}(\boldsymbol{x};
\Theta)$. ControlNet freezes $\Theta$ and instantiates a trainable copy
$\Theta_c$ of the same block, connected to the frozen path by two
\emph{zero convolutions} $\mathcal{Z}(\cdot\,;\cdot)$ --- $1\times1$ convolutions
whose weights and biases are initialised to zero. Given a conditioning vector
$\boldsymbol{c}$, the block computes
\begin{equation}
\boldsymbol{y}_c = \mathcal{F}(\boldsymbol{x}; \Theta)
+ \mathcal{Z}\!\left(\mathcal{F}\!\left(\boldsymbol{x}
+ \mathcal{Z}(\boldsymbol{c}; \Theta_{z1}); \Theta_c\right); \Theta_{z2}\right)
\label{eq:controlnet_block}
\end{equation}
Because both zero-convolution terms vanish at initialisation, the first training
step satisfies $\boldsymbol{y}_c = \boldsymbol{y}$: the augmented network is
exactly the pretrained model, and no random noise perturbs its features. The
control branch then grows its influence gradually as training proceeds, which
protects the large-scale prior inherited by the trainable copy.

The spatial condition is supplied as an image $\boldsymbol{c}_i$ at the same
resolution as the input. A lightweight encoder $\mathcal{E}_c(\cdot)$ of four
strided convolution layers maps it to the latent resolution of the backbone,
\begin{equation}
\boldsymbol{c}_f = \mathcal{E}_c(\boldsymbol{c}_i)
\label{eq:hint_encoder}
\end{equation}
and $\boldsymbol{c}_f$ is passed to the control branch. In Stable Diffusion, the
structure is replicated over the $12$ encoder blocks and the middle block of the
UNet; the resulting features are added back into the $12$ skip connections and
the middle block of the frozen decoder path. Training uses the same objective as
the backbone, extended with the spatial condition,
\begin{equation}
\mathcal{L}_{\mathrm{CN}} =
\mathbb{E}_{\boldsymbol{z}_0, t, \boldsymbol{c}_t, \boldsymbol{c}_f, \boldsymbol{\epsilon}}
\left[\left\| \boldsymbol{\epsilon} -
\boldsymbol{\epsilon}_\theta(\boldsymbol{z}_t, t, \boldsymbol{c}_t, \boldsymbol{c}_f)
\right\|_2^2\right]
\label{eq:controlnet_loss}
\end{equation}
where $\boldsymbol{c}_t$ is the text condition. Following~\cite{zhang2023adding},
a fraction of text prompts is replaced with the empty string during training,
which encourages the branch to read semantics directly from the spatial
condition rather than relying on the prompt.

\subsection{Problem Formulation}
\label{sec:formulation}

We consider the task of synthesising a thermal image directly from a
natural-language description. Let $y$ denote a text prompt and
$\boldsymbol{x}^{\mathrm{tir}} \in \mathbb{R}^{H \times W \times 3}$ the target
thermal image. Our objective is to learn a conditional distribution
$p_\theta(\boldsymbol{x}^{\mathrm{tir}} \mid y)$ from which plausible thermal
images can be sampled.

The difficulty is that this mapping is not a restyling problem. As established in
\autoref{sec:related_work}, the intensity recorded at a thermal detector is a
superposition of radiation emitted by an object at its own temperature, radiation
reflected from its surroundings, and atmospheric
radiation~\cite{vollmer2018infrared,bao2023hadar}. Under the standard
approximation $\tau_{\mathrm{atm}} \approx 1$, the signal at wavelength $\lambda$
reduces to
\begin{equation}
S_\lambda \approx e_\lambda B_\lambda(T) + (1 - e_\lambda)\,\Phi_{\mathrm{env}}
\label{eq:radiometry}
\end{equation}
where $e_\lambda$ is the surface emissivity, determined by material, $B_\lambda(T)$
is Planck's blackbody radiance at absolute temperature $T$, and
$\Phi_{\mathrm{env}}$ is the incident environmental radiation. The governing
quantities in \autoref{eq:radiometry} --- emissivity, temperature, and ambient
radiative context are \emph{not observable in the visible spectrum}. Two
vehicles identical in an RGB image differ entirely in thermal appearance
depending on whether one has been driven; a road surface at noon and at midnight
are indistinguishable in colour but inverted in the thermal.

This is precisely why RGB-to-TIR translation is
ill-posed: a single visible image is consistent with many valid thermal outputs,
and a deterministic translator must collapse that
ambiguity~\cite{Mao2026PID,Lee2026TherA}. Our formulation resolves it from the
other direction. Rather than attempting to \emph{infer} the unobservable factors,
we \emph{supply} them explicitly through language. The prompt $y$ is not a generic
scene caption but a thermally grounded description that names the attributes
appearing in \autoref{eq:radiometry}: material composition (governing
$e_\lambda$), the heat-emission state of individual objects (governing $T$), and
environmental context such as time of day and weather (governing
$\Phi_{\mathrm{env}}$). Following the caption schema
of~\cite{Lee2026TherA}, each prompt is a canonicalised
tuple
\begin{equation}
y = \{\, y_{\mathrm{scene}},\; y_{\mathrm{object}},\;
y_{\mathrm{material}},\; y_{\mathrm{heat}} \,\}
\label{eq:caption_schema}
\end{equation}
where $y_{\mathrm{scene}}$ carries the environmental context, comprising a broad
description of the setting together with the weather and time of day that determine
$\Phi_{\mathrm{env}}$; $y_{\mathrm{object}}$ identifies the objects present and
their visible attributes, including colour; $y_{\mathrm{material}}$ names the
surface composition of each object, which fixes the emissivity $e_\lambda$; and
$y_{\mathrm{heat}}$ records the heat-emission state of each object, which fixes its
absolute temperature $T$. Every token therefore carries a physically meaningful
thermal attribute rather than a description of visible appearance. In this sense the
synthesis is \emph{physics-guided}: the physical priors that a visible image cannot
carry are injected into the generator through the text channel. The individual
attributes named here --- weather, material, colour, and heat-emission state --- are
the units ablated in \autoref{tab:cn_capfield_flir}.

We instantiate two settings:
\begin{itemize}
    \item \textbf{Unconditional Text2Thermal}, which samples
    $\boldsymbol{x}^{\mathrm{tir}} \sim p_\theta(\cdot \mid y)$ from the physics-aware
    prompt alone, with no spatial input at inference.
 \item \textbf{Conditional Text2Thermal}, which additionally accepts a spatial
    condition $\boldsymbol{c}_s$, either a paired visible image or a structural map
    derived from one, and samples
    $\boldsymbol{x}^{\mathrm{tir}} \sim p_\theta(\cdot \mid y, \boldsymbol{c}_s)$
    when layout-aligned output is required.
\end{itemize}

\subsection{Text2Thermal Architecture}
\label{sec:architecture}

\subsubsection{Unconditional Text2Thermal Training}
\label{sec:uncond_training}

No thermal--caption corpus exists at a scale comparable to the web-scale
image--text datasets used to train visible-spectrum
generators~\cite{laion}, so training a thermal text-to-image model from scratch is
infeasible. We instead adapt the semantic breadth of a pretrained latent diffusion
backbone~\cite{rombach2022high} to the thermal domain, retaining its compositional
prior while retargeting its output distribution, as illustrated in \autoref{fig:sd}.

Given a thermal image $\boldsymbol{x}^{\mathrm{tir}}_0$, the frozen VAE encoder maps
it to a latent $\boldsymbol{z}_0 = \mathcal{E}(\boldsymbol{x}^{\mathrm{tir}}_0)$, to
which noise is added according to
\autoref{eq:sd_marginal}. The physics-aware prompt
$y$ of \autoref{eq:caption_schema} is encoded by the frozen CLIP text
encoder~\cite{clip} into $\tau(y)$ and injected through the UNet
cross-attention layers.

\noindent\textbf{Low-rank adaptation of the denoiser.}
Full fine-tuning of the UNet on a few thousand thermal images invites overfitting
and catastrophic forgetting of the pretrained
prior~\cite{zhang2023adding}. We therefore freeze the backbone weights and adapt
only through low-rank updates~\cite{hu2022lora}. For a frozen projection matrix
$\mathbf{W} \in \mathbb{R}^{d_{\mathrm{out}} \times d_{\mathrm{in}}}$, LoRA
introduces
\begin{equation}
\mathbf{W}' = \mathbf{W} + \Delta\mathbf{W}
= \mathbf{W} + \frac{\gamma}{r}\,\mathbf{B}\mathbf{A}
\label{eq:lora}
\end{equation}
where $\mathbf{A} \in \mathbb{R}^{r \times d_{\mathrm{in}}}$,
$\mathbf{B} \in \mathbb{R}^{d_{\mathrm{out}} \times r}$, rank
$r \ll \min(d_{\mathrm{in}}, d_{\mathrm{out}})$, and $\gamma$ is a scaling factor.
$\mathbf{B}$ is initialised to zero so that $\mathbf{W}' = \mathbf{W}$ at the start
of training and the adapted model is exactly the pretrained one.

We apply \autoref{eq:lora} to the query, key, value and output projections
$\{\mathbf{W}_Q, \mathbf{W}_K, \mathbf{W}_V, \mathbf{W}_O\}$ of \emph{both} the
self-attention and cross-attention blocks throughout the UNet encoder, middle
block, and decoder. This placement is deliberate, as the cross-attention projections are where the
text embedding $\tau(y)$ enters the network, so adapting them re-binds
RGB-trained language priors onto thermal appearance, teaching the model that
``warm engine block'' denotes a bright emissive region rather than a coloured
object, while the self-attention projections adjust the spatial statistics of
the generated latent toward the low-texture, emission-driven structure
characteristic of infrared imagery. Convolutional and normalisation layers remain frozen. With
rank $r = 16$, the trainable parameters amount to approximately $0.3\%$ of the
backbone.

Let $\theta_{\mathrm{L}} = \{\mathbf{A}_i, \mathbf{B}_i\}$ denote the LoRA
parameters. Training minimises the standard noise-prediction objective over these
parameters only,
\begin{equation}
\theta_{\mathrm{L}}^{*} = \arg\min_{\theta_{\mathrm{L}}}\;
\mathbb{E}_{\boldsymbol{z}_0, y, t, \boldsymbol{\epsilon}}
\left[\left\| \boldsymbol{\epsilon} -
\boldsymbol{\epsilon}_{\theta \oplus \theta_{\mathrm{L}}}
\!\left(\boldsymbol{z}_t, t, \tau(y)\right) \right\|_2^2\right]
\label{eq:uncond_loss}
\end{equation}
where $\theta$ are the frozen backbone weights and $\oplus$ denotes their low-rank
composition per \autoref{eq:lora}. We refer to the resulting denoiser
$\boldsymbol{\epsilon}_{\theta^{\mathrm{tir}}}$ as the \emph{thermal-adapted
backbone}; it constitutes the complete Unconditional Text2Thermal model and, as
shown next, the frozen backbone for the conditional variant.

\subsubsection{Training with Additional Control}
\label{sec:cond_training}

Language specifies \emph{what} radiates and \emph{how much}, but not \emph{where}.
A prompt describing a night-time street with active vehicles constrains the thermal
statistics of the scene without fixing object positions, extents, or boundaries.
For applications that require layout-aligned output, such as generating
synthetic thermal data against known annotations, this structural
ambiguity must be resolved by an explicit spatial signal.

We supply it through a control branch following the ControlNet
paradigm~\cite{zhang2023adding}. Two design choices distinguish our instantiation from the standard recipe, as
illustrated in \autoref{fig:cn}.

\noindent\textbf{Thermal backbone for the control branch.}
In the original ControlNet formulation, the frozen path is a vanilla RGB-pretrained
Stable Diffusion model, so the control features are added into a generator whose
output distribution is that of the visible spectrum. We instead attach the control
branch to our \emph{thermal-adapted} backbone
$\boldsymbol{\epsilon}_{\theta^{\mathrm{tir}}}$ obtained in
\autoref{sec:uncond_training}, in place of the vanilla checkpoint. The generative
prior that the frozen path contributes is therefore the thermal prior rather than
the RGB one, and the spatial signal modulates features that already live in the
thermal domain. \autoref{tab:component_ablation} quantifies the effect of this
substitution.

Formally, let $\mathcal{F}(\cdot\,;\Theta^{\mathrm{tir}})$ denote a block of the
thermal-adapted backbone and $\Theta_c$ the corresponding trainable copy in the
control branch. For a conditioning
signal $\boldsymbol{c}$, the block computes
\begin{equation}
\boldsymbol{h}_c = \mathcal{F}(\boldsymbol{h}; \Theta^{\mathrm{tir}})
+ \mathcal{Z}\!\left(\mathcal{F}\!\left(\boldsymbol{h}
+ \mathcal{Z}(\boldsymbol{c}; \Theta_{z1}); \Theta_c\right); \Theta_{z2}\right)
\label{eq:cond_block}
\end{equation}
where $\mathcal{Z}(\cdot\,;\cdot)$ are zero-initialised $1\times1$ convolutions.
Since both zero-convolution terms vanish at initialisation, the first training step
satisfies $\boldsymbol{h}_c = \mathcal{F}(\boldsymbol{h}; \Theta^{\mathrm{tir}})$:
the augmented network reproduces the Unconditional Text2Thermal model exactly, and
the visible-spectrum signal enters the thermal generator gradually rather than
perturbing it with random gradients at the outset.

\noindent\textbf{Introducing visible-spectrum structure.}
The primary spatial condition is the paired RGB frame
$\boldsymbol{c}_i = \boldsymbol{x}^{\mathrm{rgb}}$. A lightweight encoder
$\mathcal{E}_c(\cdot)$ of four strided convolution layers maps it to the latent
resolution,
\begin{equation}
\boldsymbol{c}_s = \mathcal{E}_c(\boldsymbol{x}^{\mathrm{rgb}})
\label{eq:cond_encoder}
\end{equation}
and $\boldsymbol{c}_s$ enters the branch through
\autoref{eq:cond_block}. The division of labour is explicit and, we argue, the correct one for this
modality, since the RGB image contributes only geometry, namely object
extents, boundaries, and scene layout, while the radiometric content of the
output is determined by the physics-aware prompt.

Training optimises only the control branch and the zero convolutions, with the
thermal-adapted backbone frozen:
\begin{equation}
\mathcal{L}_{\mathrm{ctrl}} =
\mathbb{E}_{\boldsymbol{z}_0, y, \boldsymbol{c}_s, t, \boldsymbol{\epsilon}}
\left[\left\| \boldsymbol{\epsilon} -
\boldsymbol{\epsilon}_{\theta^{\mathrm{tir}}}\!\left(\boldsymbol{z}_t, t,
\tau(y), \boldsymbol{c}_s\right) \right\|_2^2\right]
\label{eq:cond_loss}
\end{equation}

\noindent\textbf{Alternative spatial modalities.}
The RGB frame is one choice among several. Because paired visible imagery is not
always available, and because coarser conditions may suffice, we additionally
evaluate structural maps extracted from the RGB counterpart by large vision models:
Canny edges~\cite{canny1986}, monocular depth from Depth
Anything~\cite{yang2024depthanythingv2}, and panoptic segmentation from Grounded
SAM~\cite{ren2024grounded}. Each is substituted for
$\boldsymbol{x}^{\mathrm{rgb}}$ in \autoref{eq:cond_encoder} and a separate
branch is trained. We also evaluate composite conditions, formed by element-wise adding the
features derived from ControlNet encoders trained on different structural
modalities, before decoding, to test whether the additional cue provides
information beyond what the visible image already carries.

\subsubsection{Inference}
\label{sec:inference}
The two settings share a sampling procedure and differ only in the conditioning set.
\noindent\textbf{Unconditional Text2Thermal.}
Given a physics-aware prompt $y$, sampling begins from
$\boldsymbol{z}_T \sim \mathcal{N}(\mathbf{0}, \mathbf{I})$ and iterates the reverse
process with the thermal-adapted denoiser. Prompt adherence is controlled by
classifier-free guidance~\cite{ho2022classifierfree},
\begin{equation}
\tilde{\boldsymbol{\epsilon}} =
\boldsymbol{\epsilon}_{\theta^{\mathrm{tir}}}(\boldsymbol{z}_t, t, \varnothing)
+ s_y\left[\boldsymbol{\epsilon}_{\theta^{\mathrm{tir}}}(\boldsymbol{z}_t, t, \tau(y))
- \boldsymbol{\epsilon}_{\theta^{\mathrm{tir}}}(\boldsymbol{z}_t, t, \varnothing)\right]
\label{eq:cfg_uncond}
\end{equation}
with guidance scale $s_y$. The final latent is decoded to a thermal image,
$\hat{\boldsymbol{x}}^{\mathrm{tir}} = \mathcal{D}(\hat{\boldsymbol{z}}_0)$. No
visible-spectrum input is required at any stage, so generation is unaffected by the
low-light and adverse-weather degradation that limits translation-based
methods.
\noindent\textbf{Conditional Text2Thermal.}
When a spatial condition is available, the control branch is active and guidance is
applied over both conditions,
\begin{align}
\tilde{\boldsymbol{\epsilon}} &=
\boldsymbol{\epsilon}_{\theta^{\mathrm{tir}}}(\boldsymbol{z}_t, t, \varnothing, \varnothing) \nonumber\\
&\quad + s_s\left[\boldsymbol{\epsilon}_{\theta^{\mathrm{tir}}}(\boldsymbol{z}_t, t, \varnothing, \boldsymbol{c}_s)
- \boldsymbol{\epsilon}_{\theta^{\mathrm{tir}}}(\boldsymbol{z}_t, t, \varnothing, \varnothing)\right] \nonumber\\
&\quad + s_y\left[\boldsymbol{\epsilon}_{\theta^{\mathrm{tir}}}(\boldsymbol{z}_t, t, \tau(y), \boldsymbol{c}_s)
- \boldsymbol{\epsilon}_{\theta^{\mathrm{tir}}}(\boldsymbol{z}_t, t, \varnothing, \boldsymbol{c}_s)\right]
\label{eq:cfg_cond}
\end{align}
which first establishes scene structure $\boldsymbol{c}_s$ and then layers the
thermal semantics carried by the prompt. The scales $s_s$ and $s_y$ trade structural
adherence against radiometric expressiveness. Since the control branch is attached to a frozen
backbone, both models are served from a single set of thermal-adapted weights, and
the same prompt can be rendered with or without spatial guidance at no additional
training cost.

\subsection{Evaluating semantic fidelity}
\label{sec:eval_protocol}
Standard generative metrics do not test whether a prompt's content reaches the
output. FID compares feature distributions and is indifferent to which scene was
requested, while CLIP Score reduces both prompt and image to single embeddings and
cannot confirm that individual attributes appear. We therefore close the loop through
the captioning pipeline. Each generated image is re-captioned by the procedure of
\autoref{sec:datasets}, and the result is compared against the caption of the
corresponding ground-truth thermal frame using BERTScore~\cite{zhang2020bertscore}.
Because the captions describe material, heat-emission state, and environmental
context rather than visible appearance, agreement between the two indicates that the
physically meaningful attributes survive generation and remain recoverable from the
output.
\section{Experimental results and analysis}
\label{sec:experiments}

\subsection{Experiments setup}

\subsubsection{Datasets}
\label{sec:datasets}
All models are pretrained on R2T2~\cite{Lee2026TherA}, a corpus of $100$k
RGB--thermal--text triplets. Each triplet comprises a visible image, its aligned
thermal counterpart, and a canonicalised description of the thermal
characteristics of the scene --- object class, material, and heat-emission state,
together with environmental context such as time of day and weather.

We evaluate on three public RGB--thermal benchmarks. \textbf{FLIR}~\cite{flir} is
a driving-focused thermal dataset covering urban roads and highways under both
daytime and nighttime conditions; we use the aligned version
following~\cite{flir}. \textbf{M3FD}~\cite{liu2022target} is a
multi-scenario multi-modality benchmark spanning diverse illumination and weather
conditions. \textbf{FMB}~\cite{liu2023multi} is a full-time multi-modality
benchmark for image fusion and segmentation.
Split sizes are reported in \autoref{tab:datasets}.

\begin{table}[t]
\centering
\small
\begin{tabular}{lccc}
\hline
\textbf{Dataset} & \textbf{Train} & \textbf{Test} & \textbf{Total} \\ \hline
FLIR~\cite{flir} & 8,862 & 1,366 & 10,228 \\
M3FD~\cite{liu2022target} & 3,368 & 831 & 4,199 \\
FMB~\cite{liu2023multi} & 1,220 & 280 & 1,500 \\ \hline
\end{tabular}
\caption{Dataset splits used in our experiments. Every image is annotated with a thermal-aware caption.}\label{tab:datasets}
\end{table}

\noindent\textbf{Thermal-aware caption generation.}
None of the three benchmarks ships with textual annotations, so we generate
captions ourselves following the prompting strategy of TherA~\cite{Lee2026TherA}
without modification. A multimodal reasoning model, InternVL2.5-14B, is conditioned jointly on the
visible and thermal frames of each pair and instructed to describe how the
scene manifests in the thermal domain, producing a canonicalised schema over
scene context, object identity, material, and heat-emission state, as shown
in \autoref{eq:caption_schema}.

We report results under two protocols:
\begin{itemize}
    \item \textbf{Benchmark setting.} Models pretrained on R2T2 are subsequently
    retrained on the training split of each target benchmark and evaluated on the
    corresponding test split (\autoref{tab:main_comparison}).
    \item \textbf{Zero-shot setting.} Models are trained on R2T2 alone and
    evaluated directly on the FLIR and M3FD test splits without benchmark-specific
    adaptation, measuring how well the learned thermal prior transfers across
    sensors and scene distributions (\autoref{tab:zeroshot_comparison}).
\end{itemize}

\subsubsection{Evaluation metrics}

Text-to-thermal synthesis admits no pixel-aligned ground truth, since a
thermally grounded prompt constrains the statistics of a scene without determining a
unique image. Reference-based pixel metrics are therefore undefined in the
unconditional setting, and we evaluate along three axes.

\noindent\textbf{Distributional fidelity.} Fr\'echet Inception
Distance~\cite{heusel2017gans} (FID) measures the distance between the feature
distributions of generated and real thermal images, capturing whether the model has
learned the appearance statistics of the infrared domain.

\noindent\textbf{Text alignment.} CLIP Score reports the cosine similarity between
the embeddings of the conditioning prompt and the generated image.

\noindent\textbf{Semantic fidelity.} We additionally report \emph{reconstructed
BERTScore}, following the protocol of \autoref{sec:eval_protocol}, where each
synthesised image is re-captioned and compared against the caption of the
corresponding ground-truth thermal frame using BERTScore~\cite{zhang2020bertscore},
reported as precision, recall and F1. A high score indicates that the attributes
named in the prompt survive generation and remain recoverable from the output.

When a spatial condition is supplied the output aligns with a known layout, and we
additionally report PSNR, SSIM~\cite{wang2004image}, and
LPIPS~\cite{zhang2018unreasonable} against the paired ground-truth frame. Together
these metrics evaluate whether the generated image matches the thermal domain and
whether it reproduces the scene it was conditioned on.
\subsubsection{Training details}
All experiments are conducted on a single NVIDIA RTX A6000 (48\,GB) GPU. The
thermal-adapted backbone is obtained by low-rank adaptation of Stable
Diffusion~1.5~\cite{rombach2022high} with rank $r=16$, updating $3.19$\,M
parameters ($0.3\%$ of the backbone), while the VAE and text encoder remain
frozen. The control branch is attached to the thermal-adapted backbone and
connected through zero-initialised convolutions, with only the branch optimised.
Sampling uses 50 steps with a text guidance scale of $7.5$ and a spatial
conditioning scale of $7.5$. Complete hyperparameters, training schedules, and
per-benchmark configurations are provided in the \autoref{sec:A1}.

\subsection{Experimental results}

\begin{table}[t]
\centering
\scriptsize
\setlength{\tabcolsep}{4pt}
\begin{tabular}{llcccccc}
\hline
\multirow{2}{*}{\textbf{Method}} & \multirow{2}{*}{\makecell[c]{\textbf{Publication}\\\textbf{Venue}}} & \multicolumn{2}{c}{\textbf{M3FD}~\cite{liu2022target}} & \multicolumn{2}{c}{\textbf{FLIR}~\cite{flir}} & \multicolumn{2}{c}{\textbf{FMB}~\cite{liu2023multi}} \\ \cline{3-4} \cline{5-6} \cline{7-8}
 &  & \textbf{FID}$\downarrow$ & \makecell[c]{\textbf{CLIP}\\\textbf{Score}$\uparrow$} & \textbf{FID}$\downarrow$ & \makecell[c]{\textbf{CLIP}\\\textbf{Score}$\uparrow$} & \textbf{FID}$\downarrow$ & \makecell[c]{\textbf{CLIP}\\\textbf{Score}$\uparrow$} \\ \hline
\multicolumn{8}{l}{\textit{General image-to-image translation}} \\ \hline
InstructPix2Pix~\cite{Brooks2023InstructPix2Pix} & CVPR 2023 & 138.94 & -- & 178.03 & -- & -- & -- \\
StegoGAN~\cite{Wu_2024_CVPR} & CVPR 2024 & 247.03 & -- & 152.57 & -- & -- & -- \\
Pix2Pix-Turbo~\cite{Parmar2024OneStepIT} & arXiv 2024 & 142.26 & -- & 91.17 & -- & -- & -- \\
UNSB~\cite{Kim2024UNSB} & ICLR 2024 & 160.72 & -- & 139.46 & -- & -- & -- \\
CSGO~\cite{Xing2024CSGO} & NeurIPS 2025 & 204.53 & -- & 116.85 & -- & -- & -- \\
StyleID~\cite{Chung2024StyleID} & CVPR 2024 & 155.69 & -- & 164.43 & -- & -- & -- \\
StyleSSP~\cite{Xu2025StyleSSP} & CVPR 2025 & 165.92 & -- & 150.52 & -- & -- & -- \\
BBDM~\cite{Li2022BBDMIT} & CVPR 2023 & 238.60 & -- & 177.81 & -- & -- & -- \\ \hline
\multicolumn{8}{l}{\textit{RGB-to-TIR translation}} \\ \hline
InfraGAN~\cite{ozkanouglu2022infragan} & PRL 2022 & 254.55 & -- & 224.00 & -- & -- & -- \\
IR-Former~\cite{Chen2024ImplicitMT} & IJCNN 2024 & 196.80 & -- & 214.22 & -- & -- & -- \\
EG-GAN~\cite{lee2023edgeguided} & ICRA 2023 & 163.78 & -- & 140.92 & -- & -- & -- \\
DR-AVIT~\cite{Han2024DRAVIT} & TGRS 2024 & 161.32 & -- & 111.01 & -- & -- & -- \\
PID~\cite{Mao2026PID} & PR 2026 & 160.91 & -- & 84.26 & -- & -- & -- \\
F-ViTA~\cite{Nair2025FViTA} & WACV 2026 & 111.77 & -- & 87.03 & -- & -- & -- \\
DiffV2IR~\cite{ran2025diffv2ir} & arXiv 2025 & 92.57 & -- & 91.44 & -- & -- & -- \\
ThermalGen~\cite{Xu2025ThermalGen} & NeurIPS 2025 & 110.74 & -- & 101.45 & -- & -- & -- \\
TherA~\cite{Lee2026TherA} & CVPR 2026 & \underline{87.08} & -- & \underline{83.78} & -- & -- & -- \\ \hline
\multicolumn{8}{l}{\textit{Text-to-thermal image generation}} \\ \hline
\textbf{Text2Thermal} & \textbf{ours} & \textbf{78.18} & 0.18 & \textbf{72.67} & 0.19 & 98.11 & 0.19 \\ \hline
\end{tabular}
\caption{Comparison with existing thermal image synthesis methods on the M3FD~\cite{liu2022target}, FLIR~\cite{flir} and FMB~\cite{liu2023multi} datasets. The best results are highlighted in \textbf{bold}, and the second-best results are \underline{underlined}. The upper block reports general-purpose image-to-image translation methods and the middle block reports methods designed specifically for RGB-to-TIR translation; all of these require a paired RGB image at inference. Text2Thermal is conditioned on text, with the RGB condition additionally supplied for structural guidance.}\label{tab:main_comparison}
\end{table}

\subsubsection{Comparison with existing methods}
\label{sec:main_results}

\autoref{tab:main_comparison} reports the comparison against existing thermal
image synthesis methods on the M3FD~\cite{liu2022target},
FLIR~\cite{flir}, and FMB~\cite{liu2023multi} test sets. The upper block
lists general-purpose image-to-image translation methods and the middle block
methods designed specifically for RGB-to-TIR translation; every method in both
blocks requires a paired visible frame at inference. Text2Thermal instead generates from a physics-aware caption, with the visible
frame optionally supplying structural guidance through the control branch.

Text2Thermal achieves the best FID on the two benchmarks for which baseline results
exist. On FLIR, our model reaches $\mathbf{72.67}$, improving on the strongest prior
method, TherA~\cite{Lee2026TherA} ($83.78$), by over 11 points, and
outperforming the physics-based translator PID~\cite{Mao2026PID} by a similar
margin. On M3FD, it reaches $\mathbf{78.18}$, again ahead of TherA ($87.08$)
and improving substantially over DiffV2IR~\cite{ran2025diffv2ir} ($92.57$). We additionally present results on FMB, for which no published results exist;
our model obtains $98.11$.

CLIP Score is reported for Text2Thermal alone, as none of the compared methods
accepts a text prompt at inference. Values are stable across all three datasets
($0.18$ on M3FD, $0.19$ on FLIR and FMB), indicating that prompt adherence is
preserved as the target distribution shifts. The metric is informative for relative
comparison across configurations rather than as an absolute measure of alignment.

\begin{figure}[t]
\centering
\includegraphics[width=\linewidth]{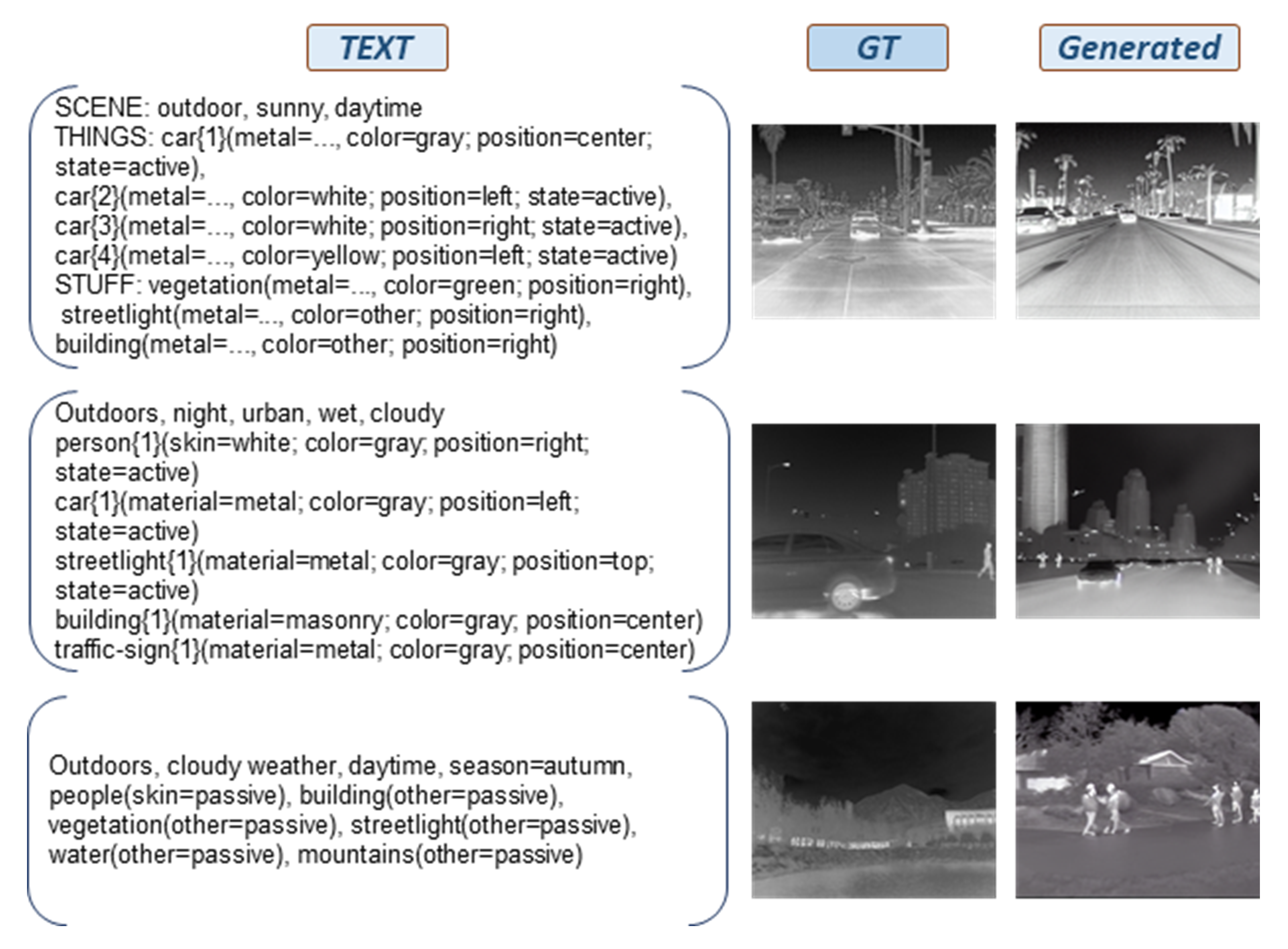}
\caption{Qualitative samples from the unconditional Text2Thermal model,
showing the input thermal text prompt alongside the ground truth and
generated thermal images.}\label{fig:sample1}
\end{figure}

\begin{figure}[t]
\centering
\includegraphics[width=\linewidth]{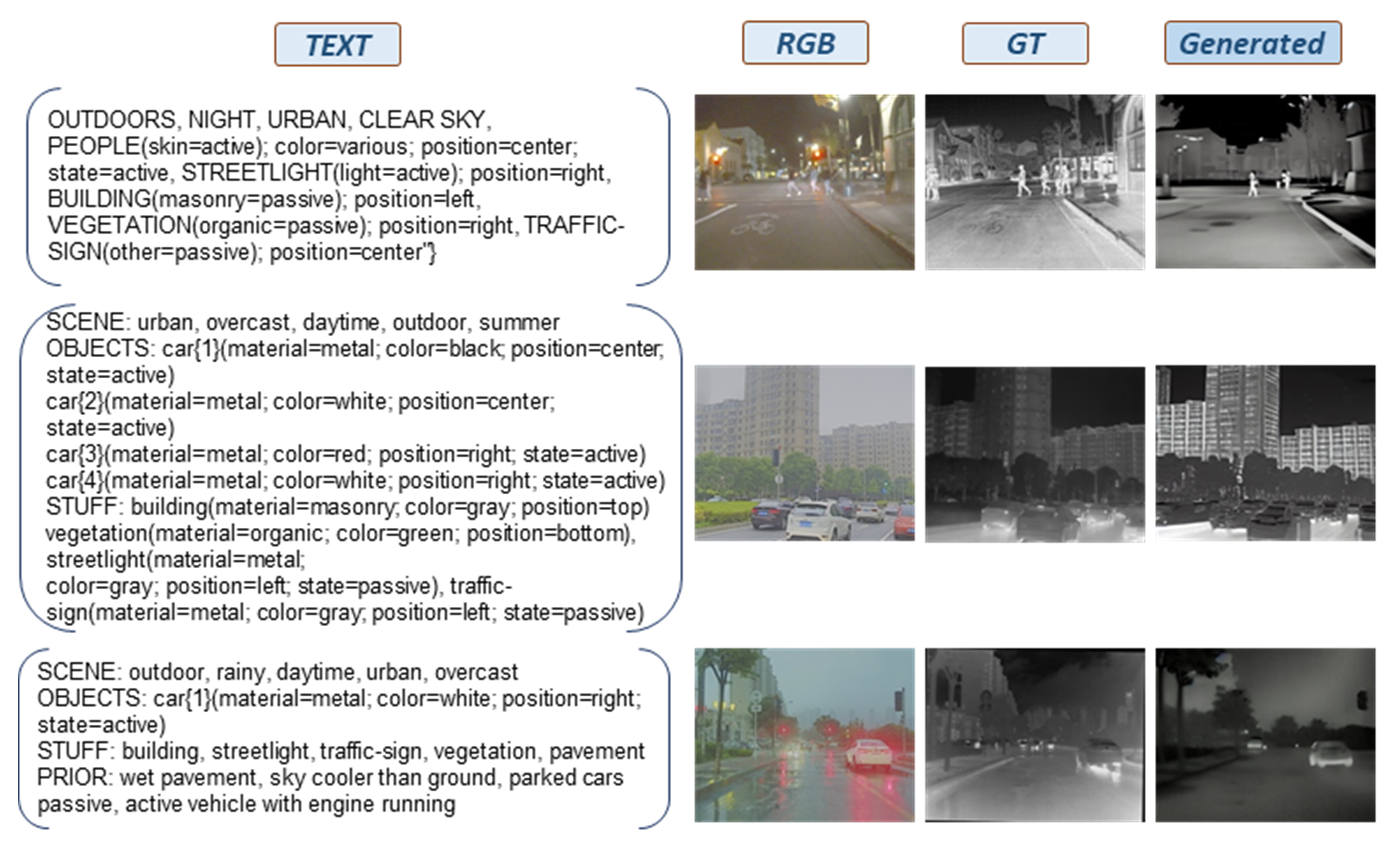}
\caption{Qualitative samples from the conditional Text2Thermal model with the
RGB frame as spatial condition, showing the input text prompt, RGB image,
ground truth thermal image, and generated output.}\label{fig:sample2}
\end{figure}

\autoref{fig:sample1} and \autoref{fig:sample2} present qualitative results
of the unconditional and conditional Text2Thermal models, with samples
across the three datasets.

\subsection{Semantic fidelity of the generated images}
\label{sec:bertscore}

\begin{table}[t]
\centering
\scriptsize
\setlength{\tabcolsep}{4pt}
\begin{tabular}{lccccccccc}
\hline
\multirow{2}{*}{\textbf{Method}} & \multicolumn{3}{c}{\textbf{M3FD}~\cite{liu2022target}} & \multicolumn{3}{c}{\textbf{FLIR}~\cite{flir}} & \multicolumn{3}{c}{\textbf{FMB}~\cite{liu2023multi}} \\ \cline{2-4} \cline{5-7} \cline{8-10}
 & \textbf{P}$\uparrow$ & \textbf{R}$\uparrow$ & \textbf{F1}$\uparrow$ & \textbf{P}$\uparrow$ & \textbf{R}$\uparrow$ & \textbf{F1}$\uparrow$ & \textbf{P}$\uparrow$ & \textbf{R}$\uparrow$ & \textbf{F1}$\uparrow$ \\ \hline
\textbf{Uncond. Text2Thermal } & 0.9071 & \textbf{0.9117} & 0.9092 & 0.8853 & 0.8856 & 0.8848 & 0.9192 & 0.9242 & 0.9216 \\
\textbf{Cond. Text2Thermal } & \textbf{0.9118} & 0.9114 & \textbf{0.9113} & \textbf{0.8959} & \textbf{0.8916} & \textbf{0.8932} & \textbf{0.9220} & 0.9242 & \textbf{0.9230} \\ \hline
\end{tabular}
\caption{Semantic fidelity of the synthesised thermal images on the M3FD~\cite{liu2022target}, FLIR~\cite{flir} and FMB~\cite{liu2023multi} datasets. The best results are highlighted in \textbf{bold}; tied values are left unmarked. Each synthesised image is re-captioned and the resulting description is compared against the caption of the corresponding ground-truth thermal image using BERTScore, reported as precision (P), recall (R) and F1. Models are pretrained on R2T2~\cite{Lee2026TherA} and subsequently retrained on each benchmark.}\label{tab:bertscore_comparison}
\end{table}

\autoref{tab:bertscore_comparison} reports reconstructed BERTScore across the three
benchmarks. Averaged over the datasets, the unconditional Text2Thermal model attains an F1 of $0.9052$
and the RGB conditional Text2Thermal $0.9092$. Scores above $0.9$ indicate close agreement
between the descriptions of generated and ground-truth thermal images, supporting the
central claim of this work: the model generates thermal content corresponding to what
the caption specifies. The results are consistent across three benchmarks.

Adding the spatial condition gives a small but consistent improvement. The margins are narrow, $0.0040$ in
average F1, which is expected: the control branch supplies geometry rather than
radiometry, and the attributes measured here are determined by the text channel in
both configurations. Semantic fidelity is therefore preserved rather than diluted
when spatial guidance is introduced.
\subsection{Ablation study}
\label{sec:ablation}
\subsubsection{Effect of Proposed Components}
\autoref{tab:component_ablation} traces the contribution of each component. The
vanilla Stable Diffusion baseline (O) reports an FID of 287.14, confirming that an RGB-pretrained generator does not transfer to
the thermal domain without adaptation. Replacing its denoiser with our
thermal-adapted UNet (A) brings FID to $109.27$, and adding the RGB condition
through the control branch (B) reduces it further to $\mathbf{72.67}$, a $36.60$-point
improvement. CLIP Score is unchanged at $0.19$ across both settings, indicating that
the structural conditioning improves distributional fidelity without altering how
faithfully the output reflects the prompt.

\begin{table}[t]
\centering
\scriptsize
\begin{tabular}{lcc}
\toprule
 & \textbf{FID}$\downarrow$ & \textbf{CLIP Score}$\uparrow$ \\ \midrule
(O) Stable Diffusion~\cite{rombach2022high} & 287.14  &0.18  \\
(A) $+$ Thermal UNet & \underline{109.27} & \underline{0.19} \\
(B) ~~~~$+$ RGB condition & \textbf{72.67} & \textbf{0.19} \\ \bottomrule
\end{tabular}
\caption{Effect of the proposed components, evaluated on the FLIR~\cite{flir} dataset. The best results are highlighted in \textbf{bold}, and the second-best results are \underline{underlined}. Starting from the original Stable Diffusion~\cite{rombach2022high} baseline (O), the proposed components are introduced one at a time, giving the incremental settings (A--B). Setting (A) replaces the visible-spectrum denoising network with our thermal-adapted UNet; setting (B) additionally supplies the paired RGB image as a second condition through the control branch, and corresponds to the complete conditional Text2Thermal model.}\label{tab:component_ablation}
\end{table}

\subsubsection{Spatial Conditioning Modality}

\begin{figure}[t]
\centering
\includegraphics[width=\linewidth]{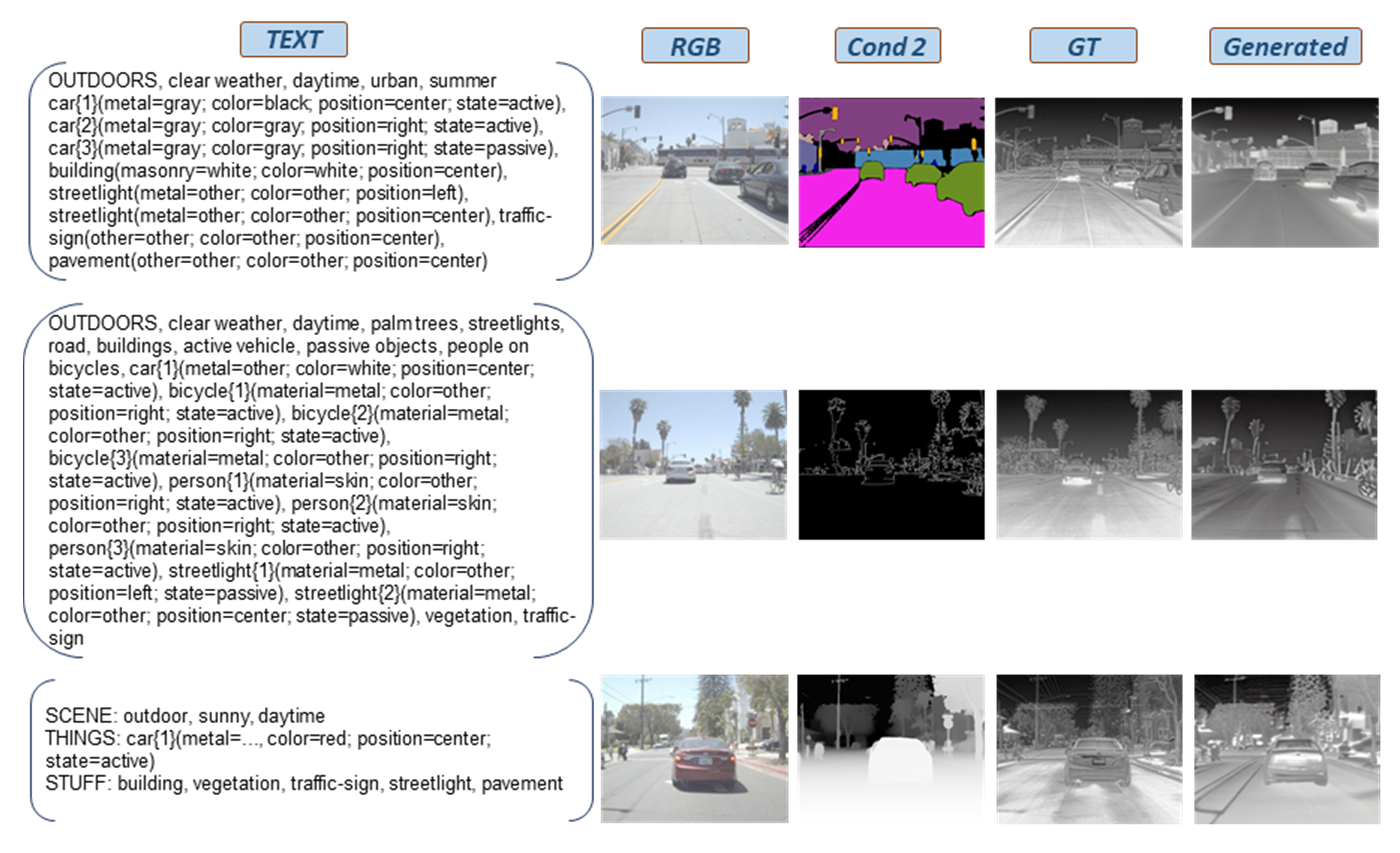}
\caption{Qualitative examples of the spatial conditioning modalities
(segmentation, edge, and depth maps) used alongside the RGB frame,
with corresponding thermal captions, ground truth, and generated outputs.}\label{fig:sample3}
\end{figure}

\autoref{tab:cn_modality_flir} compares spatial conditioning modalities, and
\autoref{fig:sample3} presents the corresponding samples. The RGB frame is the
strongest single condition, with an FID of $\mathbf{72.67}$
against $99.30$--$105.74$ for edge, depth, and segmentation maps, indicating that the
extracted structural abstractions discard information the visible image retains.
Composite conditions do not help; every RGB-plus-structure combination degrades FID
substantially (best case $93.90$ for RGB\,$+$\,Depth),
while improving SSIM only marginally. The structural information available to the
model saturates with the RGB frame alone, and additional maps introduce conflicting
guidance rather than complementary detail.
\subsubsection{Which Caption Fields Carry Thermal Information?}
\autoref{tab:cn_capfield_flir} removes one attribute class from the caption schema at
a time. Every removal degrades FID, by $12.58$ to $16.90$ points, confirming that all
four classes carry thermal information. Material is the
most costly to remove ($126.17$), consistent with its direct role in
\autoref{eq:radiometry} through emissivity, followed by weather ($124.54$). Object state contributes least ($121.85$), but its removal still degrades FID
by $12.58$ points, and the margins separating each class are narrow enough
that the ranking should be read as indicative rather than decisive.

\begin{table}[t]
\centering
\scriptsize
\begin{tabular}{lccccc}
\hline
\textbf{Spatial Condition} & \textbf{SSIM}$\uparrow$ & \textbf{PSNR}$\uparrow$ & \textbf{LPIPS}$\downarrow$ & \textbf{FID}$\downarrow$ & \textbf{CLIP Score}$\uparrow$ \\ \hline
Edge & 0.3624 & 12.33 & 0.5288 & 100.00 & 0.1865 \\
Depth & 0.3715 & 13.27 & 0.5236 & 99.30 & 0.1863 \\
Segmentation map & 0.3682 & 11.79 & 0.5346 & 105.74 & 0.1863 \\
RGB image (Text2Thermal) & 0.4043 & \textbf{14.22} & \textbf{0.4855} & \textbf{72.67} & \textbf{0.1919} \\ \hline
RGB $+$ Edge & \textbf{0.4089} & \underline{13.50} & 0.4980 & 96.33 & 0.1889 \\
RGB $+$ Depth & 0.4048 & 13.46 & \underline{0.4954} & \underline{93.90} & 0.1887 \\
RGB $+$ Seg. & \underline{0.4088} & 13.10 & 0.5044 & 97.40 & \underline{0.1894} \\ \hline
\end{tabular}
\caption{Ablation on the spatial conditioning modality $c_2$, evaluated on the FLIR \cite{flir} dataset. The best results are highlighted in \textbf{bold}, and the second-best results are \underline{underlined}. All configurations retain the text condition $c_1$ and differ only in the spatial signal supplied to the control branch. The lower block reports composite conditions, testing whether a third modality yields further gains beyond the RGB image or whether the structural information saturates.}\label{tab:cn_modality_flir}
\end{table}

\begin{table}[t]
\centering
\scriptsize
\begin{tabular}{lccccc}
\hline
\textbf{Configuration} & \textbf{Weather} & \textbf{Material} & \textbf{Color} & \textbf{State} & \textbf{FID}$\downarrow$ \\ \hline
Full schema (Text2Thermal) & \checkmark & \checkmark & \checkmark & \checkmark & \textbf{109.27} \\
\hline
w/o Weather (99.9\%) & $\times$ & \checkmark & \checkmark & \checkmark & 124.54 \\
w/o Material (87.1\%) & \checkmark & $\times$ & \checkmark & \checkmark & 126.17 \\
w/o Color (83.9\%) & \checkmark & \checkmark & $\times$ & \checkmark & 121.98 \\
w/o State (97.0\%) & \checkmark & \checkmark & \checkmark & $\times$ & \underline{121.85} \\ \hline
\end{tabular}
\caption{Leave-one-attribute-out ablation over the structured caption schema, evaluated on the FLIR \cite{flir} dataset on Unconditional Text2Thermal model. The attributes correspond to the fields of \autoref{eq:caption_schema}: weather is carried by $y_{\mathrm{scene}}$, colour by $y_{\mathrm{object}}$, material by $y_{\mathrm{material}}$, and heat-emission state by $y_{\mathrm{heat}}$. The best results are highlighted in \textbf{bold}, and the second-best results are \underline{underlined}. Each row removes a single attribute class from the text condition at both training and inference time; the resulting degradation indicates how much thermal information that class carries. Percentages give the fraction of training captions actually modified by each removal.}\label{tab:cn_capfield_flir}
\end{table}

\section{Discussion}
\label{sec:discussion}

\subsection{Parameter efficiency}
\label{sec:discussion-1}
\autoref{tab:efficiency} compares trainable and total parameter counts against
diffusion-based thermal synthesis methods on FLIR. Because our approach adapts a
pretrained stable diffusion model, the unconditional Text2Thermal model updates only
$\mathbf{3.19}$\,M parameters --- $1.4\%$ of the $225.53$\,M optimised by
PID~\cite{Mao2026PID} and TherA~\cite{Lee2026TherA} while reaching an FID of
$109.27$. Adding the control branch brings $361.28$\,M trainable parameters and
the best FID in the table ($\mathbf{72.67}$), a significant improvement over
TherA at a comparable training cost. The total parameter count is higher for both
of our configurations, since the frozen Stable Diffusion backbone is retained in
full; this affects memory at inference but not training cost, and the two
configurations share a single set of backbone weights, so the conditional model
adds only the control branch on top of the unconditional Text2Thermal one rather than requiring a
separate deployment.

\begin{table}[t]
\centering
\scriptsize
\setlength{\tabcolsep}{6pt}
\begin{tabular}{lccc}
\hline
\textbf{Method} & \makecell[c]{\textbf{Trainable}\\\textbf{Param. (M)}$\downarrow$} & \makecell[c]{\textbf{Total}\\\textbf{Param. (M)}$\downarrow$} & \textbf{FID}$\downarrow$ \\ \hline
PID~\cite{Mao2026PID}   & 225.53 & \textbf{309.18} & 84.26 \\
TherA~\cite{Lee2026TherA} & 225.53 & \underline{309.18} & \underline{83.78} \\ \hline
Uncond. Text2Thermal       & \textbf{3.19}   & 1066.37 & 109.27 \\
Cond Text2Thermal & \underline{361.28} & 1427.65 & \textbf{72.67} \\ \hline
\end{tabular}
\caption{Parameter efficiency comparison against diffusion-based thermal synthesis methods, with FID reported on the FLIR test set. The best results are highlighted in \textbf{bold}, and the second-best results are \underline{underlined}.}\label{tab:efficiency}
\end{table}

\subsection{Zero-Shot Generalization}
\label{sec:discussion-2}

\begin{table}[t]
\centering
\scriptsize
\setlength{\tabcolsep}{5pt}
\begin{tabular}{llcccc}
\hline
\multirow{2}{*}{\textbf{Method}} & \multirow{2}{*}{\makecell[c]{\textbf{Publication}\\\textbf{Venue}}} & \multicolumn{2}{c}{\textbf{M3FD}~\cite{liu2022target}} & \multicolumn{2}{c}{\textbf{FLIR}~\cite{flir}} \\ \cline{3-4} \cline{5-6}
 &  & \textbf{FID}$\downarrow$ & \makecell[c]{\textbf{CLIP}\\\textbf{Score}$\uparrow$} & \textbf{FID}$\downarrow$ & \makecell[c]{\textbf{CLIP}\\\textbf{Score}$\uparrow$} \\ \hline
F-ViTA~\cite{Nair2025FViTA} & WACV 2026 & 145.83 & -- & 145.83 & -- \\
DiffV2IR~\cite{ran2025diffv2ir} & arXiv 2025 & 132.49 & -- & 253.82 & -- \\
ThermalGen~\cite{Xu2025ThermalGen} & NeurIPS 2025 & 177.85 & -- & 127.14 & -- \\
TherA~\cite{Lee2026TherA} & CVPR 2026 & \underline{105.52} & -- & \textbf{112.93} & -- \\ \hline
\textbf{Text2Thermal } & ours & \textbf{104.50} & 0.1793 & \underline{125.56} & 0.1779 \\ \hline
\end{tabular}
\caption{Zero-shot evaluation on the M3FD \cite{liu2022target} and FLIR \cite{flir} datasets. The best results are highlighted in \textbf{bold}, and the second-best results are \underline{underlined}. All models are trained on R2T2~\cite{Lee2026TherA} without dataset-specific fine-tuning, and evaluated directly on each benchmark. }\label{tab:zeroshot_comparison}
\end{table}

\autoref{tab:zeroshot_comparison} evaluates transfer without dataset-specific
adaptation, where all models are trained on R2T2~\cite{Lee2026TherA} and applied
directly to the M3FD and FLIR test sets.
Text2Thermal obtains the best FID on M3FD, marginally ahead of
TherA~\cite{Lee2026TherA}, and the second-best on FLIR. Both results improve
substantially on the remaining baselines. CLIP Scores of $\approx 0.18$ are close to the fine-tuned values, indicating that prompt adherence
is largely retained even where distributional fidelity drops. A
text-conditioned generator remaining competitive with translation methods
under zero-shot transfer is notable, given that those methods receive a
paired visible frame at inference on every test image, since the thermal
prior learned from physics-aware captions evidently encodes attributes
relevant to the thermal domain.
\subsection{Failure Cases}

\begin{figure}[t]
\centering
\includegraphics[width=\linewidth]{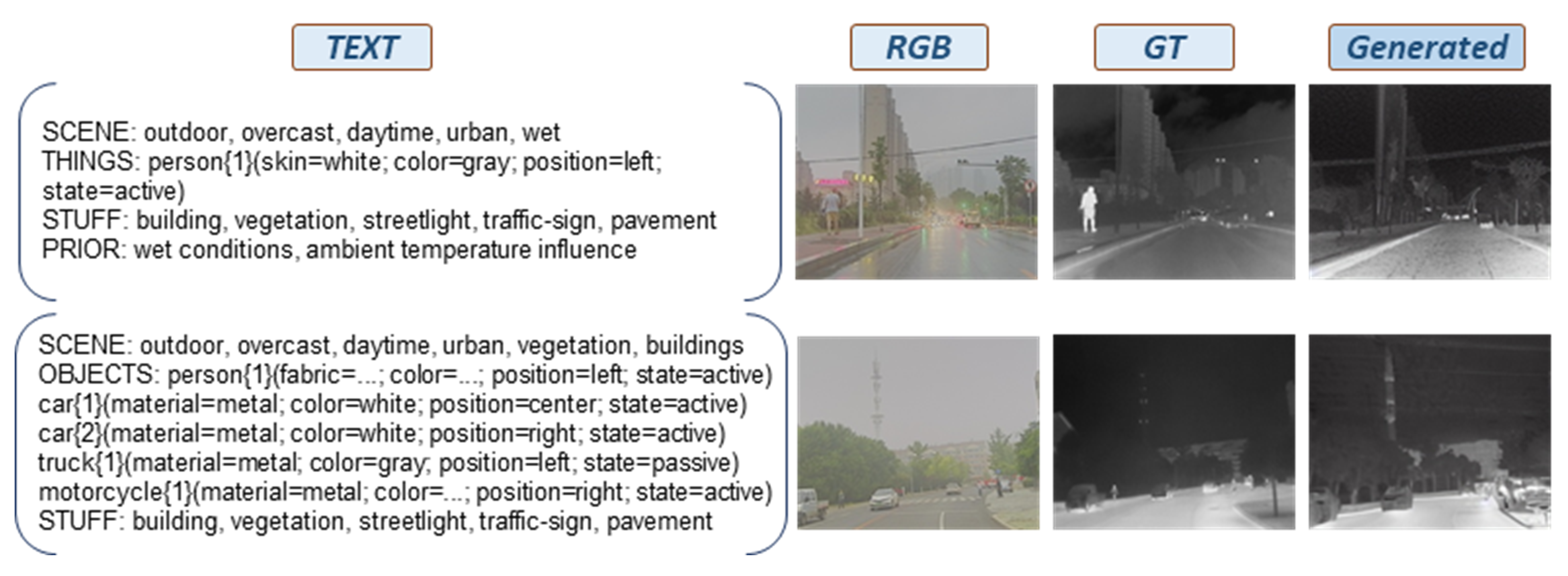}
\caption{Failure cases of Text2Thermal. In the first row, the generated
thermal image omits the pedestrian present in the ground truth despite being
specified in the text prompt. In the second row, the model hallucinates a
tower structure that is not clearly visible in the ground truth thermal
image, illustrating the spurious generation of structural elements absent from
the reference.}\label{fig:sample4}
\end{figure}

\autoref{fig:sample4} presents some samples of failure cases, where the
generated thermal image either omits objects named in the prompt, such as the
pedestrian in the first row, or hallucinates structures carried over from the
spatial condition, such as the tower in the second row.
\section{Conclusion}
\label{sec:conclusion}
We presented Text2Thermal, a framework for physics-aware thermal image synthesis that
supplies the unobservable thermal properties through language rather than inferring
them from RGB. Our experiments establish four findings. Thermal images can be
generated from text alone by adapting a large-scale text-to-image backbone to the
thermal domain, and across M3FD and FLIR the model achieves the best FID among
reported thermal image synthesis methods while updating only a small fraction of the
backbone parameters. We also report thermal synthesis results on the FMB dataset.
Re-captioning the generated images and comparing against the ground-truth captions
with BERTScore confirms that the synthesised imagery carries the thermal content the
prompt specified. Removing any single attribute class from the prompt degrades
generation quality, with material the most costly, consistent with its role in
setting emissivity. This confirms that the prompt functions as a thermal
physics-aware specification rather than a stylistic cue. Finally, the RGB frame
supplies complete structural information for the task, since adding edge, depth, or
segmentation maps alongside it degrades rather than improves generation quality.

We note several limitations of the present framework, each suggesting a direction for
future work. The physics encoded in our captions is soft, since the descriptions are
produced by prompting a multimodal model on paired visible and thermal frames, so the
model learns a linguistic proxy for emissivity and temperature rather than the
quantities themselves. Calibrated radiometric supervision could be explored to
replace this proxy with an explicit physical constraint. Constructing the captions
requires paired RGB and TIR data, so methods that generate effective captions from
thermal images alone, without a visible counterpart, would open the framework to the
many thermal datasets that carry no aligned RGB imagery. The generated images also
sometimes omit objects named in the prompt or hallucinate structures absent from the
reference, suggesting a need for finer object-level control.

More broadly, the framework lowers the barrier to working in the infrared domain.
Thermal training data can be obtained without a thermal sensor, the prompt gives
direct control over the conditions generated, and generation quality is better than
that of translation-based methods.

\section*{CRediT authorship contribution statement}

\textbf{Tayeba Qazi:} Conceptualization, Methodology, Software, Validation,
Formal analysis, Investigation, Data curation, Writing -- original draft,
Visualization. \textbf{Brejesh Lall:} Writing -- review \& editing,
Supervision, Project administration. \textbf{Prerana Mukherjee:} Writing --
review \& editing, Supervision.

\section*{Declaration of competing interest}

The authors declare that they have no known competing financial interests or
personal relationships that could have appeared to influence the work reported
in this paper.

\section*{Declaration of generative AI and AI-assisted technologies in the writing process}

During the preparation of this work the authors used Claude (Anthropic) in
order to assist with language editing and formatting of the manuscript. After
using this tool, the authors reviewed and edited the content as needed and take
full responsibility for the content of the published article.

\section*{Acknowledgements}

The authors gratefully acknowledge the support of the Prime Minister's Research
Fellowship (PMRF), Ministry of Education, Government of India (PMRF ID:
1402111), for funding this research.

\section*{Data availability}

All datasets used in this work are publicly available. All other data will be
made available on request.

\bibliographystyle{elsarticle-num}
\bibliography{references}

\appendix
\renewcommand{\thetable}{A.\arabic{table}}
\renewcommand{\thefigure}{A.\arabic{figure}}
\setcounter{figure}{0}
\setcounter{table}{0}

\section{Additional Implementation Details}
\label{sec:A1}

Implementation and training details for both settings are summarised in
\autoref{tab:impl}. Entries marked --- do not apply to that setting.

\begin{table}[t]
\centering
\scriptsize
\setlength{\tabcolsep}{4pt}
\begin{tabular}{lll}
\toprule
& \textbf{Unconditional} & \textbf{Conditional} \\
\midrule
\multicolumn{3}{c}{\textbf{Training}} \\
\midrule
Backbone & Stable Diffusion~1.5 & Stable Diffusion~1.5 + LoRA-fused UNet \\
Conditioning branch & --- & ControlNet, full fine-tune \\
ControlNet init. & --- & Thermal-adapted (LoRA-fused) SD1.5 UNet encoder \\
Text encoder & \multicolumn{2}{l}{Long-CLIP-L (frozen), 248 tokens} \\
Initialisation & R2T2-pretrained LoRA & --- \\
LoRA rank $r$ / $\alpha$ & 16 / 16 (scaling $\gamma = 1.0$) & --- \\
LoRA targets & $W_q, W_k, W_v, W_o$, all attention & --- \\
Conditioning input & --- & RGB image, $256\times256$ \\
Empty-prompt proportion & --- & 0.5 (classifier-free guidance) \\
Trainable parameters & 3.19\,M ($0.3\%$) & ControlNet encoder ($\sim$361\,M) \\
Frozen & --- & VAE, thermal-adapted UNet, text encoder \\
Optimiser & \multicolumn{2}{l}{AdamW ($\beta_1{=}0.9$, $\beta_2{=}0.999$, $\epsilon{=}10^{-8}$)} \\
Weight decay & \multicolumn{2}{l}{$10^{-2}$} \\
Learning rate & $2\times10^{-5}$, cosine, 100 warm-up & $1\times10^{-5}$, constant, 500 warm-up \\
Gradient clipping & \multicolumn{2}{l}{1.0} \\
Gradient checkpointing & --- & Enabled \\
Batch size & \multicolumn{2}{l}{16 $\times$ 2 accum. (effective 32)} \\
Training steps & 2{,}000 & 25{,}000 \\
Resolution & \multicolumn{2}{l}{$256\times256$, centre crop} \\
Precision & \multicolumn{2}{l}{fp16 mixed precision} \\
Noise schedule & \multicolumn{2}{l}{DDPM, 1{,}000 steps, scaled-linear} \\
Prediction target & \multicolumn{2}{l}{$\epsilon$} \\
Checkpointing & --- & Every 1{,}000 steps, last 3 kept \\
Seed & \multicolumn{2}{l}{42} \\
\midrule
\multicolumn{3}{c}{\textbf{Inference}} \\
\midrule
Sampler & PNDM, 50 steps & UniPC, 50 steps \\
Text guidance scale $s_y$ & \multicolumn{2}{l}{7.5} \\
ControlNet cond. scale & --- & 1.0 \\
Conditioning input & --- & RGB image, $256\times256$ \\
Negative prompt & \multicolumn{2}{l}{$\varnothing$ (empty string)} \\
Resolution & \multicolumn{2}{l}{$256\times256$} \\
Precision & \multicolumn{2}{l}{fp16} \\
Seed & \multicolumn{2}{l}{42} \\
\bottomrule
\end{tabular}
\caption{Implementation details for the unconditional and conditional Text2Thermal models. Settings shared by both are spanned across the two columns.}\label{tab:impl}
\end{table}

%
\end{document}